\documentclass[conference]{IEEEtran}
\IEEEoverridecommandlockouts

\usepackage{cite}
\usepackage{amsmath,amssymb,amsfonts}
\usepackage{algorithmic}
\usepackage{graphicx}
\usepackage{textcomp}
\usepackage{xcolor}
\def\BibTeX{{\rm B\kern-.05em{\sc i\kern-.025em b}\kern-.08em
    T\kern-.1667em\lower.7ex\hbox{E}\kern-.125emX}}
\usepackage{amsmath,graphicx,hyperref}
\usepackage{booktabs} 
\usepackage{array}    
\usepackage{enumitem}
\setlist[itemize]{leftmargin=*}

\makeatletter
\newcommand{\linebreakand}{%
  \end{@IEEEauthorhalign}
  \hfill\mbox{}\par
  \mbox{}\hfill\begin{@IEEEauthorhalign}
}
\makeatother

\begin{document}

\title{Beyond Description: A Multimodal Agent Framework for Insightful Chart Summarization}

\author{\IEEEauthorblockN{Yuhang Bai}
\IEEEauthorblockA{\textit{The Hong Kong Polytechnic University}} 
\and
\IEEEauthorblockN{Yujuan Ding}
\IEEEauthorblockA{\textit{The Hong Kong Polytechnic University}  }
\and
\IEEEauthorblockN{Shanru Lin}
\IEEEauthorblockA{\textit{City University of Hong Kong}  }
\linebreakand
\IEEEauthorblockN{Wenqi Fan}
\IEEEauthorblockA{\textit{The Hong Kong Polytechnic University} }

}

\maketitle

\begin{abstract}
Chart summarization is crucial for enhancing data accessibility and the efficient consumption of information. However, existing methods, including those with Multimodal Large Language Models (MLLMs), primarily focus on low-level data descriptions and often fail to capture the deeper insights which are the fundamental purpose of data visualization. To address this challenge, we propose Chart Insight Agent Flow, a plan-and-execute multi-agent framework effectively leveraging the perceptual and reasoning capabilities of MLLMs to uncover profound insights directly from chart images.  Furthermore, to overcome the lack of suitable benchmarks, we introduce ChartSummInsights, a new dataset featuring a diverse collection of real-world charts paired with high-quality, insightful summaries authored by human data analysis experts. Experimental results demonstrate that our method significantly improves the performance of MLLMs on the chart summarization task, producing summaries with deep and diverse insights.

\end{abstract}

\section{Introduction}

\begin{figure*}[h]
    \centering
    \includegraphics[width=0.85\linewidth]{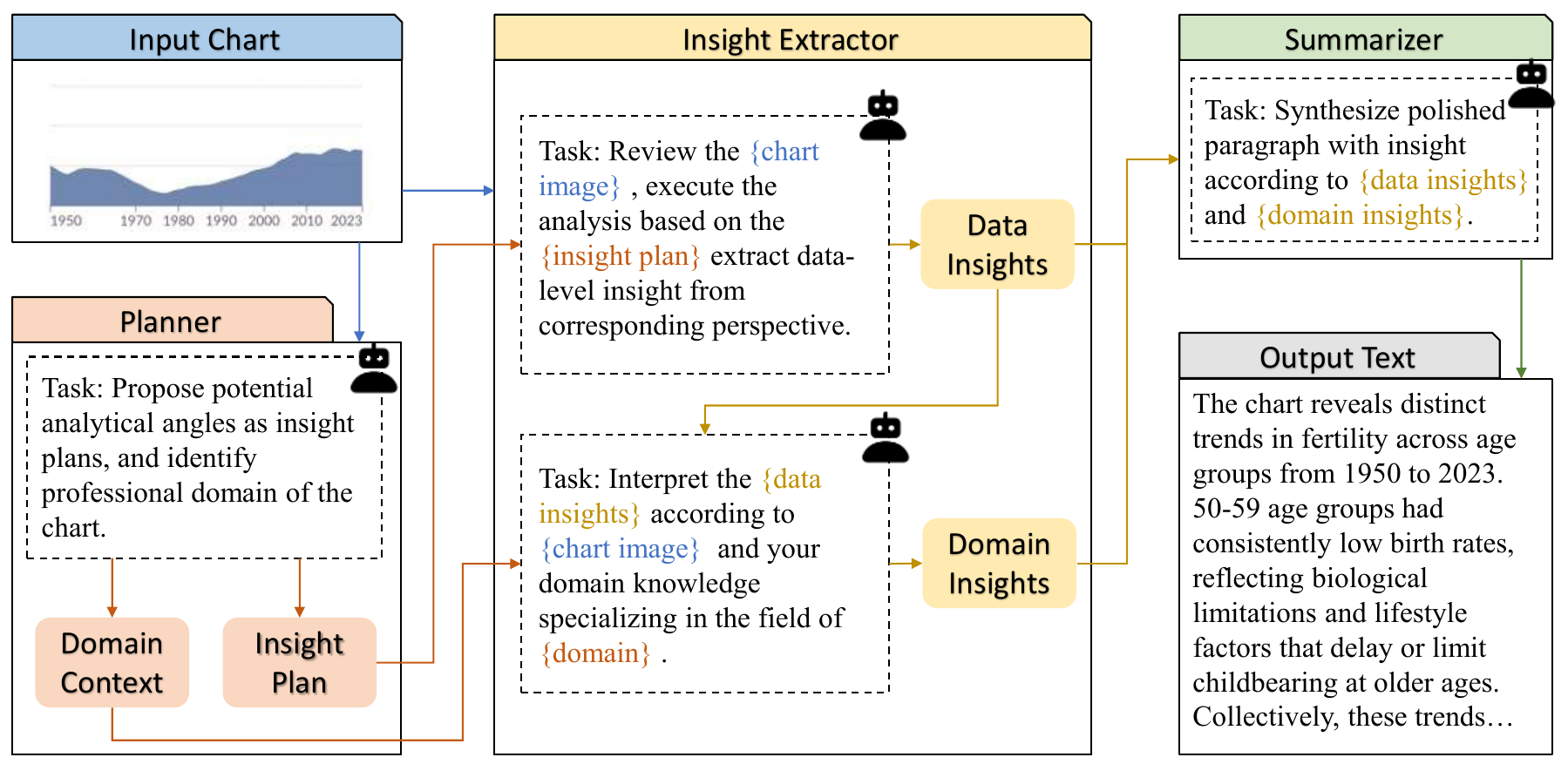}
    \caption{Chart Insight Agent Flow (CIAF) Framework, which contains three core components: Planner, Insight Extractor and Summarizer.}
    \label{fig:framework}
\end{figure*}

Data visualization, particularly through charts, is an indispensable tool for efficiently communicating complex information, making chart analysis a significant research topic~\cite{masry2024chartinstruct}. A key task in this field, chart summarization, aims to automatically generate a natural language summary that describes a chart's core content and findings~\cite{rahman2023chartsumm,masry2023unichart}. This capability is essential for enhancing data accessibility, enabling the rapid consumption of large volumes of data, and facilitating data-driven storytelling~\cite{wang2021survey}. Ultimately, it helps users quickly grasp key information, greatly enhancing efficiency and addressing information overload~\cite{chang2009defining,card1999readings,kantharaj2022chart}. To achieve these goals, chart summarization methods must effectively bridge the gap between visual data and natural language, which is a challenging yet significant research area~\cite{he2024leveraging}.

The emergence of powerful Multimodal Large Language Models (MLLMs) has introduced new paradigms for this task by demonstrating remarkable abilities in visual understanding and reasoning. While many existing methods have made progress in describing basic visual elements and factual data, they remain limited to low-level analysis~\cite{ko2024natural,sultanum2023datatales,tang2023vistext} such as semantic understanding~\cite{zhang2025mllms}, often overlooking the most critical aspect of visualization: insight~\cite{card1999readings,pousman2007casual}. The true purpose of a visualization is not merely to display data but to convey the deeper insights hidden within it~\cite{choe2015characterizing}. These insights include high-level patterns, trends, and domain-related impacts~\cite{he2024leveraging,choe2015characterizing}. Effectively summarizing a chart, therefore, requires uncovering these underlying insights with accurate language. However, extracting such insights remains a considerable challenge. It necessitates deep analysis and a solid understanding of both the presented data and the underlying domain knowledge~\cite{battle2023exactly,liu2014effects}. Despite these challenges, several recent studies have made attempts. Some methods leverage Large Language Models (LLMs) to extract insights from raw structured data, but they fail to interpret visual information~\cite{ding2023insightpilot, sahu2024insightbench, weng2025insightlens}. Other works like ChartInsights~\cite{wu2024chartinsights} and ChartInsighter~\cite{wang2025chartinsighter} focus on multimodal understanding but are limited to low-level ChartQA tasks or specific chart types, and often require access to the raw data.

To fill this research gap, we propose a novel plan-and-execute multi-agent framework, the Chart Insight Agent Flow (CIAF), which is a training-free pipeline based on MLLMs. As illustrated in Fig. ~\ref{fig:framework}, the CIAF decomposes the complex task of generating insightful chart summaries into three specialized stages. These stages are processed sequentially by a Planner Agent, an Insight Extraction Agent, and a Summarizer Agent, with each agent responsible for a distinct part of the process, from preliminary planning to final summary generation.
A significant obstacle to progress in this field has been the lack of a suitable benchmark. Existing datasets for insight generation are typically derived from raw data tables~\cite{sahu2024insightbench}, lacking the visual modality of charts. Conversely, chart-specific datasets often focus on low-level tasks visual question answering and summarization~\cite{masry2022chartqa, rahman2023chartsumm, kantharaj2022chart}. We contribute the ChartSummInsights dataset, which is uniquely constructed from real-world chart images and corresponding summaries annotated by human experts to capture profound, domain-specific insights. This expert-level annotation provides a reliable ground truth and aligns with the rich prior knowledge of MLLMs, enhancing their ability to generate coherent and contextually accurate analyses.

Our contributions are threefold: 1) We contribute the ChartSummInsights dataset, a unique collection of multi-type charts with expert-crafted, insightful summaries;
2) We propose the Chart Insight Agent Flow (CIAF), a method that leverages MLLMs' capabilities to generate insightful summaries for charts; and
3) We design a new evaluation mechanism to assess models based on the quality and diversity of insights, providing a valid measure for this task.

\section{Method}

In this paper, we introduce a straightforward yet effective agent flow for generating insightful chart summaries, which we call the Chart Insight Agent Flow (CIAF). We leverage the capabilities of state-of-the-art MLLMs to understand visual data and interpret semantic information, as well as reasoning. Rather than building a new MLLM from scratch with complex structures or optimization techniques, our approach focuses on a novel agent-based methodology that effectively utilizes existing models to accomplish a specific task: interpreting and summarizing key insights from a given chart in clear natural language. Therefore, our method is feasible to be applied on different MLLM backbones. We do not aim to improve the chart interpretation capabilities of visual models or the sentence generation quality of language models. Instead, our primary goal is to produce chart summaries with higher-quality insights. To achieve this, our framework, as shown in Fig.~\ref{fig:framework}, consists of three core agent components—Planner, Insight Extractor, and Summarizer—each responsible for distinct tasks.

\noindent \textbf{Planner.}
The Planner serves as the initial stage for interpreting the input chart image, which accepts chart images and performs two key functions:

\textit{Insight Plan Generation}: The Planner analyzes the visual and semantic elements of the chart image and proposes a set of potential analytical perspectives as an insight plan. This plan outlines a sequence of analytical actions to guide subsequent insight extraction process. We use examples of possible insight perspectives paired with according chart type and specific professional domain as an In Context Learning (ICL) prompt, which facilitate MLLM learns and imitates the structured approach to insight planning.

\textit{Domain Identification}: The Planner identifies the professional domain most relevant to the chart, which is used to prompt the following Insight Extraction Module, enabling more specialized and knowledge-grounded analysis.

The output of the Planner is a structured plan containing both a list of insight plan and a domain label, which are passed to the Insight Extraction Module for execution.

\noindent \textbf{Insight Extractor.}
Based on theoretical research on Insights~\cite{battle2023exactly,chen2009toward}, two key sources to produce insights from a chart are data and relevant domain behind it. Therefore, we device two independent agents in this stage to collaboratively extract more comprehensive insights, serving as Data Analyst and Domain Analyst respectively: 

\textit{Data Analyst}: It follows the insight plan generated by the Planner and extracts data insights from the chart. An ICL prompt containing multiple examples of data insight are provided. Insights extracted in this module are used for further in-depth analysis of subsequent modules.

\begin{figure}
    \centering
    \includegraphics[width=\linewidth]{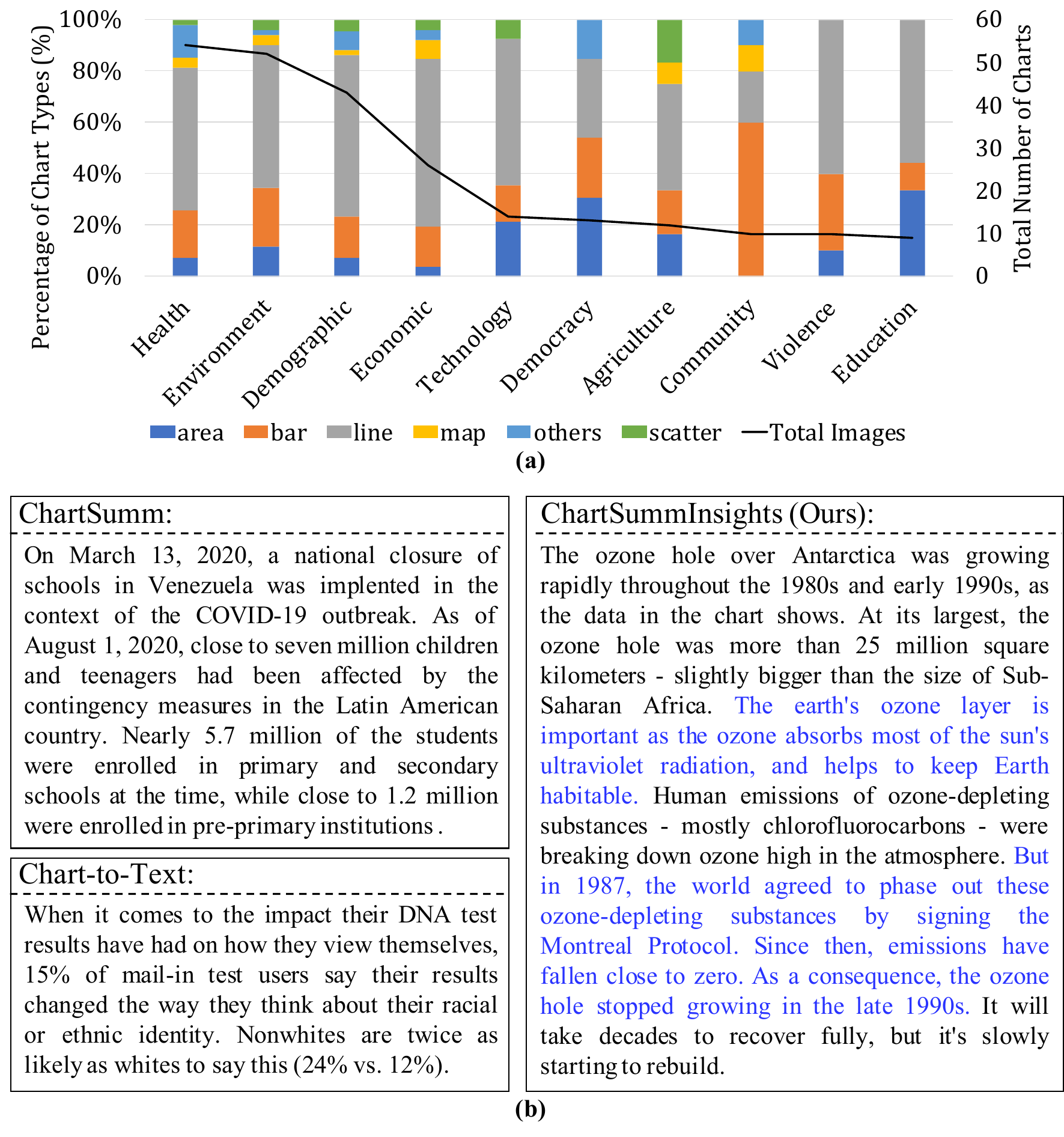}
    \caption{ChartSummInsight dataset. (a) Chart Type and Domain Distribution; (b) Sample comparison with existing chart summarization datasets~\cite{kantharaj2022chart,rahman2023chartsumm}. } 
    \label{fig:dataexample}
\end{figure}

\textit{Domain Analyst}: It is prompted with domain context provided by the Planner and plays the role of a domain expert. It takes the data insights produced by the Data Analyst and enhances them with relevant background knowledge, contextual implications, and real-world significance. 

The result is a set of domain insights that are both meaningful and actionable within the identified domain.

The two agent analysts complement each other, thereby producing two types of insights: data and domain, which together form a comprehensive analytical view, ensuring diverse and complete insights to be extracted in this part. 

\noindent \textbf{Summarizer.}
The Summarizer is implemented with an LLM, which functions to integrate the data and domain insights into a fluent, logically structured summary. The model is prompted to synthesize all extracted information, and produce a coherent narrative that captures both statistical and domain-specific aspects of the chart. The final output is a well-rounded summarization that is not only factually accurate but also contextually relevant and rich in insight.

\begin{figure}
    \centering
    \includegraphics[width=\linewidth]{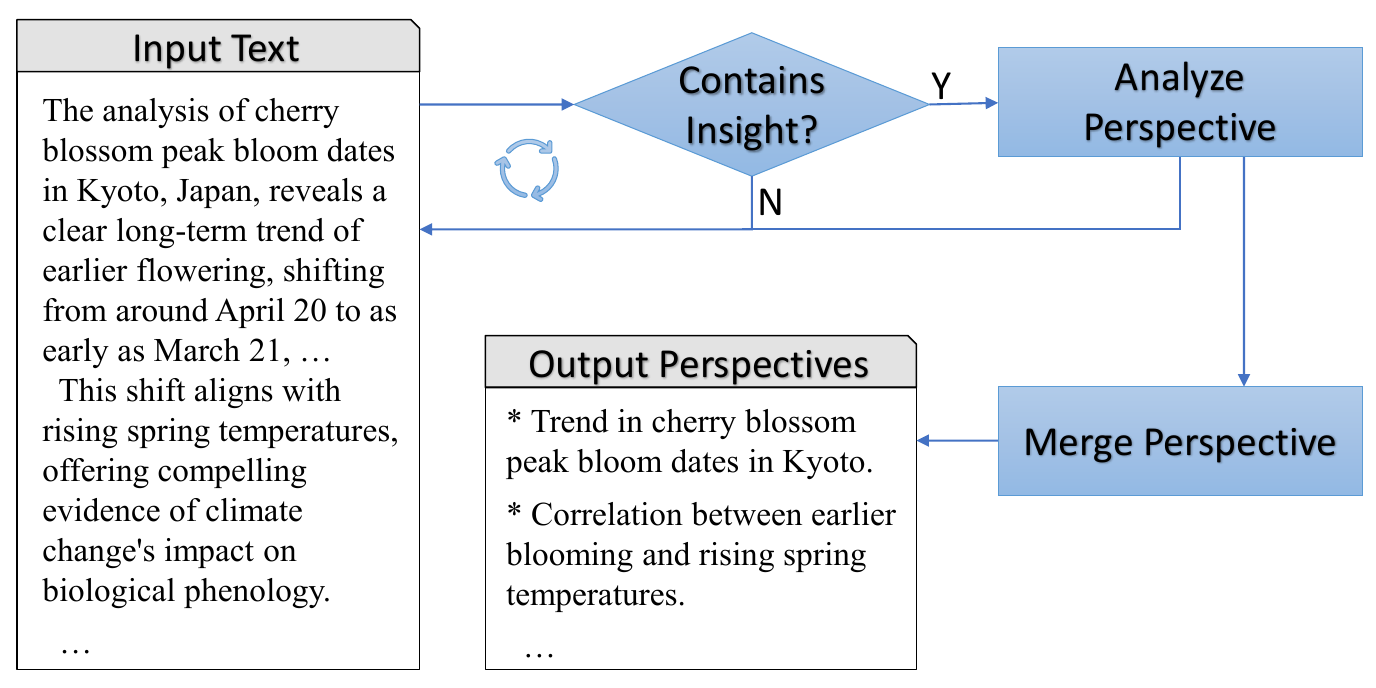}
    \caption{Insight Perspective Analysis Process}
    \label{fig:dataprocess}
\end{figure}

\begin{table}[htbp]
  \centering
  \caption{Chart Insight Summarization Performance}\vspace{2mm}
  \label{tab:experiment}
  \begin{tabular}{cccc}
    \toprule
    Baselines & ID-RC 
    & ID-Span & IQ Score \\
    \midrule
    Unichart~\cite{masry2023unichart} & 0.27 
    & 0.27 & 1.22 \\
    Matcha~\cite{liu2022matcha} & 0.11 
    & 0.11 & 1.16 \\
    \midrule 
    ChartLlama~\cite{han2023chartllama}  & 0.32 
    &	0.32 & 1.91 \\
    ChartInstruct~\cite{masry2024chartinstruct}  & 0.45 
    & 0.44 & 2.25 \\
    ChartGemma~\cite{masry2024chartgemma}  & 0.38 
    & 0.37 & 2.71 \\
    \midrule 
    QwenVL-3b~\cite{wang2024qwen2}  & 0.54 
    & 0.53 & 3.68 \\
    QwenVL-7b~\cite{wang2024qwen2}  & 0.54 
    & 0.52 & 4.09 \\
    QwenVL-plus~\cite{wang2024qwen2}  & 0.62 
    & 0.60 & 4.14 \\
    LlaVa-7b~\cite{liu2023visual}  & \textbf{0.65} 
    & \textbf{0.63} & 2.29 \\
    InternVL-2b~\cite{chen2024internvl}  & 0.54 
    & 0.53 &  3.25 \\
    InternVL-8b~\cite{chen2024internvl}  & 0.64 
    & 0.62 & 3.97 \\
    \midrule 
    QwenVL-plus-ours~\cite{wang2024qwen2}  & \textbf{0.65}
    & \textbf{0.63} & \textbf{4.67} \\
    \bottomrule
  \end{tabular}

\end{table}

\section{Experiments}
\textbf{Dataset.}
We collected a chart insight dataset of 240 images from Our World in Data\footnote{https://ourworldindata.org/}, which is a platform producing charts regarding the problems faced by the world, as well as the insightful elaboration produced by researchers at the University of Oxford and the non-profit organization Global Change Data Lab. In our collected dataset, each data point consists of a data visualization chart paired with a corresponding, expertly written summarization with insight.

The dataset spans multiple professional domains and a variety of common chart types. As shown in Fig.~\ref{fig:dataexample}, Compared with other chart summarization datasets that simply describe the data information in the chart, our dataset contains more in-depth and actionable insights, which can better evaluate the model's visual perception capabilities across different chart types and its knowledge reasoning abilities within various professional domains.

\noindent \textbf{Baselines.} 
After surveying previous chart-to-text generation approaches, we selected the following models from 3 categories for comparison, including Early-stage end-to-end chart-to-text models (Unichart~\cite{masry2023unichart}, Matcha~\cite{liu2022matcha}); Fine-tuned MLLM-based moedls (ChartLlama~\cite{han2023chartllama}, ChartInstruct~\cite{masry2024chartinstruct} and ChartGemma~\cite{masry2024chartgemma}); Off-the-shelf MLLMs (Qwen-VL~\cite{wang2024qwen2}, LLaVa~\cite{liu2023visual} and Intern-VL~\cite{chen2024internvl} series).

\begin{table}[htbp]
\setlength{\tabcolsep}{2.5pt}
  \centering
  \caption{Performance on different backbone LLMs}\vspace{2mm}
  \label{tab:experiment_backbone}
  \resizebox{\linewidth}{!}{
  \begin{tabular}{ccccccc}
    \toprule
     &QwenVL &QwenVL &QwenVL &LlaVa &InternVL &InternVL \\
    ~ &-3b-ours &-7b-ours &-plus-ours &-7b-ours &-2b-ours &-8b-ours \\ \midrule 
     ID-RC &0.56 & 0.61 & 0.65 & 0.69 & 0.64 & 0.69\\
      (+)  &0.02 & 0.07 & 0.03 & 0.04 & 0.10 & 0.05 \\
      (+\%)  &3.70 & 12.96 & 4.84 & 6.15 & 18.51 & 7.81\\ \midrule 
     ID-Span & 0.54 & 0.60 & 0.63 & 0.66 & 0.61 & 0.67  \\
      (+)  & 0.01 & 0.08 & 0.03 & 0.03 & 0.08 & 0.05\\
       (+\%) & 1.89 & 15.38 & 5.00 & 4.76 & 15.10 & 8.06\\ \midrule 
    IQ Score &4.14 & 4.33 &4.67 & 2.51 & 3.98 & 4.56\\ 
       (+)     & 0.46 & 0.24 & 0.53 & 0.22 & 0.73 & 0.59\\
       (+\%)     & 12.50 & 5.87 &12.80 & 9.61 & 22.46 & 14.86 \\

    \bottomrule
  \end{tabular}
  }
\end{table}

\begin{figure}[h]
    \centering
    \includegraphics[width=\linewidth]{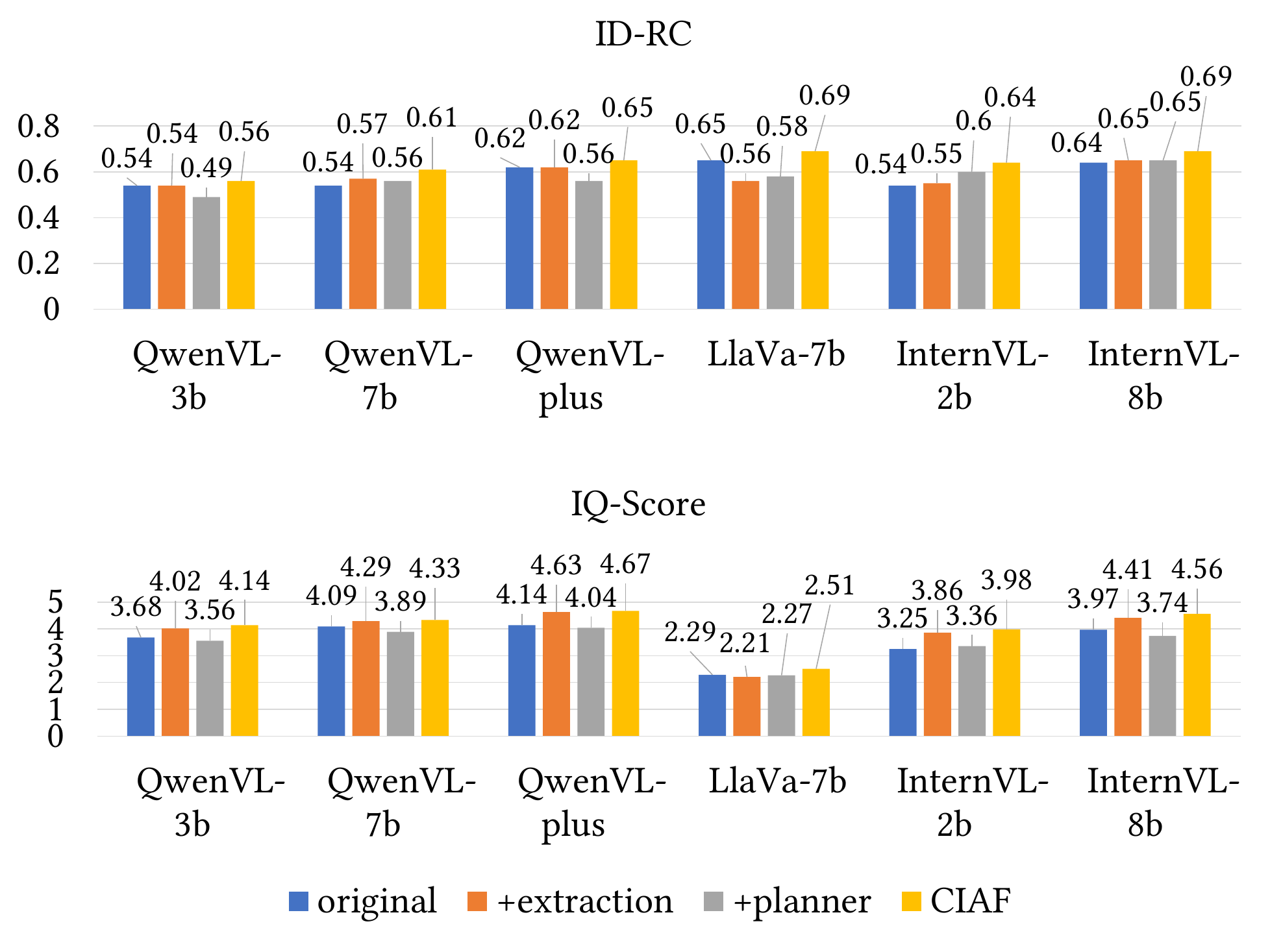}
    \caption{Performance of models with different agent components on various backbone models.}
    \label{fig:ablation}
\end{figure}

\noindent \textbf{Summarized Insight Evaluation}. 
To assess the performance of various baseline models, we designed a two-pronged evaluation framework focused specifically on the insight level, rather than just the overall summary. This approach allows us to measure both the \textbf{quality} and \textbf{diversity} of the generated content. For Insight Quality (IQ) Score, we apply an LLM (GPT) to measure the depth and factual correctness of the generated insights. Each output is given a score from 1 to 5 based on how well it aligns with the ground-truth insights and facts. This method provides a nuanced measure of the model's ability to produce meaningful and accurate observations.
For Insight Diversity (\textbf{ID}), we employ multiple BERT-based metrics~\cite{ko2024natural} including remote-clique (\textbf{RC}) and \textbf{Span}, which measure different aspects of the variety within the generated insights. To facilitate this evaluation, we first use a separate LLM agent (GPT) to post-process the generated summary. As shown in Fig.~\ref{fig:dataprocess}, this agent analyzes each sentence in the summary to extract distinct insight perspectives. These extracted perspectives are then used to calculate the diversity metrics, ensuring our evaluation is based on the core ideas, not just the surface-level wording.

\noindent \textbf{Experimental Results.}
Experimental results are shown in Table~\ref{tab:experiment}. Our approach effectively enhances both the GPT score and the SBERT-based diversity score of the generated insight summarization across the tested backbone models. Furthermore, our method performance surpasses all fine-tuned MLLM-based methods and Early end-to-end models compared. This indicates that our method not only produces summaries with greater depth and accuracy but also achieves more comprehensive analysis, leading to more insightful summarization with rich content.

To further analyze the performance of each module, we conducted ablation study on each core component. As shown in Fig.~\ref{fig:ablation}, without insight planning, the extraction module can generate in-depth insights, while lacking diversity. Thus, insight planner has been proved to successfully enrich the insight generation perspective. The overall score significantly decreased after the removal of the insight extraction module, demonstrating its effectiveness of executing the insight plan.

\begin{figure}[h]
    \centering
    \includegraphics[width=0.9\linewidth]{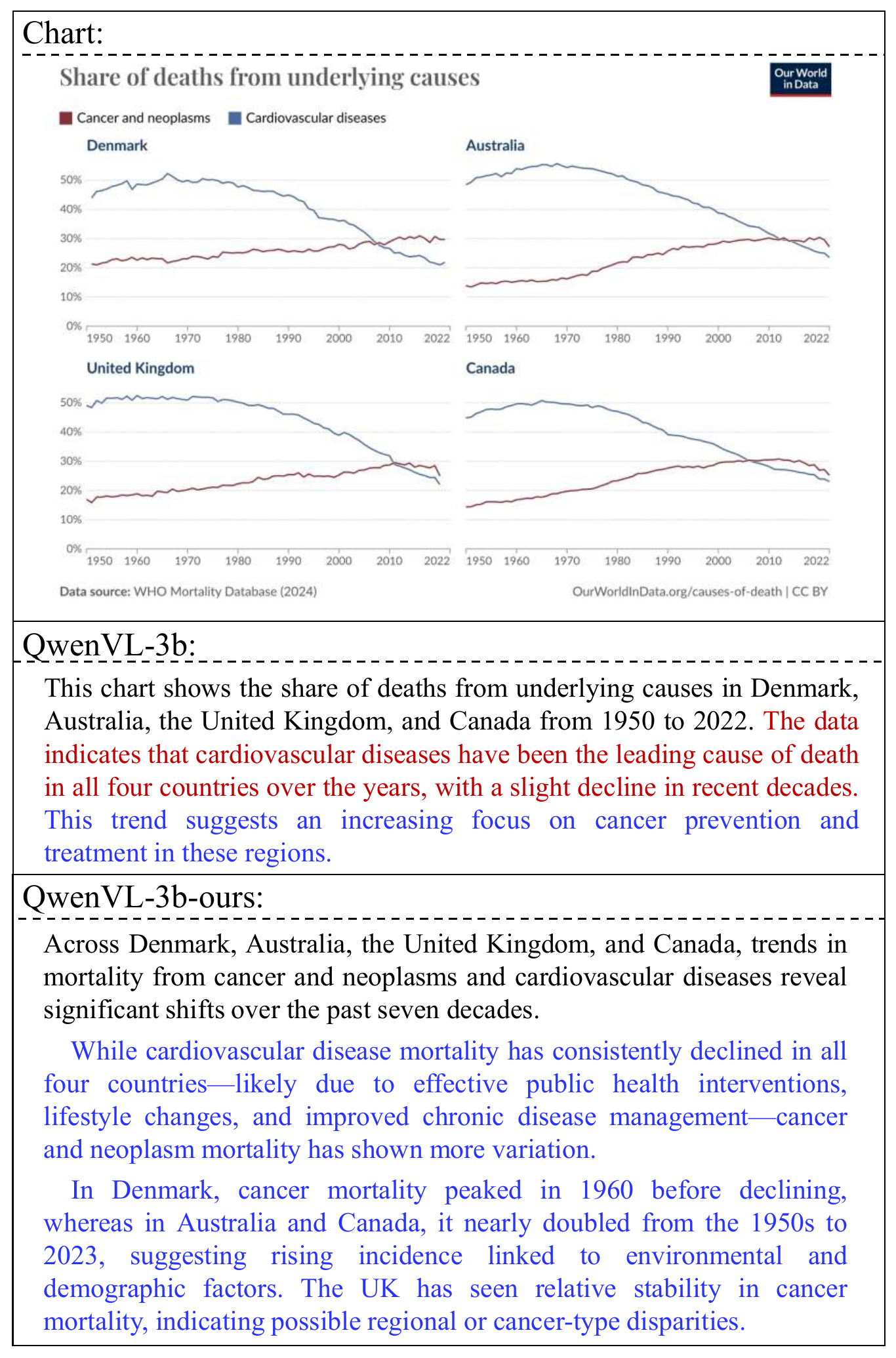}
    \caption{A comparison case showing summary generation results of our proposed method and compared QwenVL-3b.}
    \label{fig:case}
\end{figure}
\noindent \textbf{Case Study.}
Fig.~\ref{fig:case} shows an output case generated by our model, compared with output of off-the-shelf MLLM backbones. The text colored in black indicates sentences that lack insight and merely state facts, while the text colored in blue indicates insightful chart analysis. The text colored in red indicates incorrect facts. These results indicate that our approach can generate chart summaries with domain-relevant insights more effectively, surpassing the baseline model in analytical depth. At the same time, our method significantly reduces the factual errors caused by LLMs' hallucinations.

\section{Conclusion}
In this work, we propose a plan-and-execute multi-agent framework Chart Insight Agent Flow based on MLLMs, which can facilitate the chart summarization task with insightful findings. Additionally, a new dataset ChartSummInsights is collected with real-world charts annotated by human data analysis experts as a benchmark. Our method shows improved performance, demonstrating its ability to improve insight in chart summarization tasks.

\bibliographystyle{IEEEbib}
\footnotesize
\bibliography{refs}

\begin{thebibliography}{10}

\bibitem{masry2024chartinstruct}
Ahmed Masry, Mehrad Shahmohammadi, Md~Rizwan Parvez, Enamul Hoque, and Shafiq Joty,
\newblock ``Chartinstruct: Instruction tuning for chart comprehension and reasoning,''
\newblock {\em arXiv preprint arXiv:2403.09028}, 2024.

\bibitem{rahman2023chartsumm}
Raian Rahman, Rizvi Hasan, Abdullah~Al Farhad, Md~Tahmid~Rahman Laskar, Md~Hamjajul Ashmafee, and Abu Raihan~Mostofa Kamal,
\newblock ``Chartsumm: A comprehensive benchmark for automatic chart summarization of long and short summaries,''
\newblock {\em arXiv preprint arXiv:2304.13620}, 2023.

\bibitem{masry2023unichart}
Ahmed Masry, Parsa Kavehzadeh, Xuan~Long Do, Enamul Hoque, and Shafiq Joty,
\newblock ``Unichart: A universal vision-language pretrained model for chart comprehension and reasoning,''
\newblock {\em arXiv preprint arXiv:2305.14761}, 2023.

\bibitem{wang2021survey}
Qianwen Wang, Zhutian Chen, Yong Wang, and Huamin Qu,
\newblock ``A survey on ml4vis: Applying machine learning advances to data visualization,''
\newblock {\em IEEE transactions on visualization and computer graphics}, vol. 28, no. 12, pp. 5134--5153, 2021.

\bibitem{chang2009defining}
Remco Chang, Caroline Ziemkiewicz, Tera~Marie Green, and William Ribarsky,
\newblock ``Defining insight for visual analytics,''
\newblock {\em IEEE Computer Graphics and Applications}, vol. 29, no. 2, pp. 14--17, 2009.

\bibitem{card1999readings}
Stuart~K Card, Jock Mackinlay, and Ben Shneiderman,
\newblock {\em Readings in information visualization: using vision to think},
\newblock Morgan Kaufmann, 1999.

\bibitem{kantharaj2022chart}
Shankar Kantharaj, Rixie Tiffany~Ko Leong, Xiang Lin, Ahmed Masry, Megh Thakkar, Enamul Hoque, and Shafiq Joty,
\newblock ``Chart-to-text: A large-scale benchmark for chart summarization,''
\newblock {\em arXiv preprint arXiv:2203.06486}, 2022.

\bibitem{he2024leveraging}
Yi~He, Shixiong Cao, Yang Shi, Qing Chen, Ke~Xu, and Nan Cao,
\newblock ``Leveraging foundation models for crafting narrative visualization: A survey,''
\newblock {\em arXiv preprint arXiv:2401.14010}, 2024.

\bibitem{ko2024natural}
Hyung-Kwon Ko, Hyeon Jeon, Gwanmo Park, Dae~Hyun Kim, Nam~Wook Kim, Juho Kim, and Jinwook Seo,
\newblock ``Natural language dataset generation framework for visualizations powered by large language models,''
\newblock in {\em Proceedings of the 2024 CHI Conference on Human Factors in Computing Systems}, 2024, pp. 1--22.

\bibitem{sultanum2023datatales}
Nicole Sultanum and Arjun Srinivasan,
\newblock ``Datatales: Investigating the use of large language models for authoring data-driven articles,''
\newblock in {\em 2023 IEEE Visualization and Visual Analytics (VIS)}. IEEE, 2023, pp. 231--235.

\bibitem{tang2023vistext}
Benny~J Tang, Angie Boggust, and Arvind Satyanarayan,
\newblock ``Vistext: A benchmark for semantically rich chart captioning,''
\newblock {\em arXiv preprint arXiv:2307.05356}, 2023.

\bibitem{zhang2025mllms}
Xiao Zhang, Dongyuan Li, Liuyu Xiang, Yao Zhang, Cheng Zhong, and Zhaofeng He,
\newblock ``Do mllms really understand the charts?,''
\newblock {\em arXiv preprint arXiv:2509.04457}, 2025.

\bibitem{pousman2007casual}
Zachary Pousman, John Stasko, and Michael Mateas,
\newblock ``Casual information visualization: Depictions of data in everyday life,''
\newblock {\em IEEE transactions on visualization and computer graphics}, vol. 13, no. 6, pp. 1145--1152, 2007.

\bibitem{choe2015characterizing}
Eun~Kyoung Choe, Bongshin Lee, et~al.,
\newblock ``Characterizing visualization insights from quantified selfers' personal data presentations,''
\newblock {\em IEEE computer graphics and applications}, vol. 35, no. 4, pp. 28--37, 2015.

\bibitem{battle2023exactly}
Leilani Battle and Alvitta Ottley,
\newblock ``What exactly is an insight? a literature review,''
\newblock {\em 2023 IEEE Visualization and Visual Analytics (VIS)}, pp. 91--95, 2023.

\bibitem{liu2014effects}
Zhicheng Liu and Jeffrey Heer,
\newblock ``The effects of interactive latency on exploratory visual analysis,''
\newblock {\em IEEE transactions on visualization and computer graphics}, vol. 20, no. 12, pp. 2122--2131, 2014.

\bibitem{ding2023insightpilot}
R~Ding, S~Han, and D~Zhang,
\newblock ``Insightpilot: An llm-empowered automated data exploration system,''
\newblock in {\em EMNLP 2023. ACL special interest group on linguistic data (SIGDAT)}, 2023.

\bibitem{sahu2024insightbench}
Gaurav Sahu, Abhay Puri, Juan Rodriguez, Amirhossein Abaskohi, Mohammad Chegini, Alexandre Drouin, Perouz Taslakian, Valentina Zantedeschi, Alexandre Lacoste, David Vazquez, et~al.,
\newblock ``Insightbench: Evaluating business analytics agents through multi-step insight generation,''
\newblock {\em arXiv preprint arXiv:2407.06423}, 2024.

\bibitem{weng2025insightlens}
Luoxuan Weng, Xingbo Wang, Junyu Lu, Yingchaojie Feng, Yihan Liu, Haozhe Feng, Danqing Huang, and Wei Chen,
\newblock ``Insightlens: Augmenting llm-powered data analysis with interactive insight management and navigation,''
\newblock {\em IEEE Transactions on Visualization and Computer Graphics}, 2025.

\bibitem{wu2024chartinsights}
Yifan Wu, Lutao Yan, Leixian Shen, Yunhai Wang, Nan Tang, and Yuyu Luo,
\newblock ``Chartinsights: Evaluating multimodal large language models for low-level chart question answering,''
\newblock {\em arXiv preprint arXiv:2405.07001}, 2024.

\bibitem{wang2025chartinsighter}
Fen Wang, Bomiao Wang, Xueli Shu, Zhen Liu, Zekai Shao, Chao Liu, and Siming Chen,
\newblock ``Chartinsighter: An approach for mitigating hallucination in time-series chart summary generation with a benchmark dataset,''
\newblock {\em IEEE Transactions on Visualization and Computer Graphics}, 2025.

\bibitem{masry2022chartqa}
Ahmed Masry, Do~Xuan Long, Jia~Qing Tan, Shafiq Joty, and Enamul Hoque,
\newblock ``Chartqa: A benchmark for question answering about charts with visual and logical reasoning,''
\newblock {\em arXiv preprint arXiv:2203.10244}, 2022.

\bibitem{chen2009toward}
Yang Chen, Jing Yang, and William Ribarsky,
\newblock ``Toward effective insight management in visual analytics systems,''
\newblock in {\em 2009 IEEE Pacific Visualization Symposium}. IEEE, 2009, pp. 49--56.

\bibitem{liu2022matcha}
Fangyu Liu, Francesco Piccinno, Syrine Krichene, Chenxi Pang, Kenton Lee, Mandar Joshi, Yasemin Altun, Nigel Collier, and Julian~Martin Eisenschlos,
\newblock ``Matcha: Enhancing visual language pretraining with math reasoning and chart derendering,''
\newblock {\em arXiv preprint arXiv:2212.09662}, 2022.

\bibitem{han2023chartllama}
Yucheng Han, Chi Zhang, Xin Chen, Xu~Yang, Zhibin Wang, Gang Yu, Bin Fu, and Hanwang Zhang,
\newblock ``Chartllama: A multimodal llm for chart understanding and generation,''
\newblock {\em arXiv preprint arXiv:2311.16483}, 2023.

\bibitem{masry2024chartgemma}
Ahmed Masry, Megh Thakkar, Aayush Bajaj, Aaryaman Kartha, Enamul Hoque, and Shafiq Joty,
\newblock ``Chartgemma: Visual instruction-tuning for chart reasoning in the wild,''
\newblock {\em arXiv preprint arXiv:2407.04172}, 2024.

\bibitem{wang2024qwen2}
Peng Wang, Shuai Bai, Sinan Tan, Shijie Wang, Zhihao Fan, Jinze Bai, Keqin Chen, Xuejing Liu, Jialin Wang, Wenbin Ge, et~al.,
\newblock ``Qwen2-vl: Enhancing vision-language model's perception of the world at any resolution,''
\newblock {\em arXiv preprint arXiv:2409.12191}, 2024.

\bibitem{liu2023visual}
Haotian Liu, Chunyuan Li, Qingyang Wu, and Yong~Jae Lee,
\newblock ``Visual instruction tuning,''
\newblock {\em Advances in neural information processing systems}, vol. 36, pp. 34892--34916, 2023.

\bibitem{chen2024internvl}
Zhe Chen, Jiannan Wu, Wenhai Wang, Weijie Su, Guo Chen, Sen Xing, Muyan Zhong, Qinglong Zhang, Xizhou Zhu, Lewei Lu, et~al.,
\newblock ``Internvl: Scaling up vision foundation models and aligning for generic visual-linguistic tasks,''
\newblock in {\em Proceedings of the IEEE/CVF conference on computer vision and pattern recognition}, 2024, pp. 24185--24198.

\end{thebibliography}

\end{document}